\documentclass[journal]{IEEEtran}
\usepackage{cite}

\ifCLASSINFOpdf
\else
   \usepackage[dvips]{graphicx}
   \graphicspath{{../eps/}}
   \DeclareGraphicsExtensions{.eps}
\fi
\usepackage[tbtags]{amsmath}
\usepackage{mathtools}
\usepackage{amsfonts}
\usepackage{amssymb}

\usepackage{algorithmic}
\usepackage{algorithm}
\usepackage{multirow}               

\usepackage{booktabs}

\usepackage{array}
\begin{document}
%
\title{Simultaneously Learning Neighborship and\\Projection Matrix for Supervised\\Dimensionality Reduction}
%
%
%

\author{Yanwei~Pang,~\IEEEmembership{Senior~Member,~IEEE,}
	Bo~Zhou,~
	and~Feiping~Nie,~\IEEEmembership{Senior~Member,~IEEE}
\thanks{Y. Pang and B. Zhou are with the School of Electrical and Information Engineering, Tianjin University, Tianjin 300072, China (e-mail: pyw@tju.edu.cn; zhoubo@tju.edu.cn).}
\thanks{F. Nie is with the Center for OPTical IMagery Analysis and Learning (OPTIMAL), Northwestern Polytechnical University, Xi'an, 710072, Shaanxi, China (e-mail: feipingnie@gmail.com).}}

%
%

\markboth{arXiv version}
{Shell \MakeLowercase{\textit{et al.}}: Bare Demo of IEEEtran.cls for IEEE Journals}
%



\maketitle

\begin{abstract}
Explicitly or implicitly, most of dimensionality reduction methods need to determine which samples are neighbors and the similarity between the neighbors in the original high-dimensional space. The projection matrix is then learned on the assumption that the neighborhood information (e.g., the similarity) is known and fixed prior to learning. However, it is difficult to precisely measure the intrinsic similarity of samples in high-dimensional space because of the curse of dimensionality. Consequently, the neighbors selected according to such similarity might and the projection matrix obtained according to such similarity and neighbors are not optimal in the sense of classification and generalization. To overcome the drawbacks, in this paper we propose to let the similarity and neighbors be variables and model them in low-dimensional space. Both the optimal similarity and projection matrix are obtained by minimizing a unified objective function. Nonnegative and sum-to-one constraints on the similarity are adopted. Instead of empirically setting the regularization parameter, we treat it as a variable to be optimized. It is interesting that the optimal regularization parameter is adaptive to the neighbors in low-dimensional space and has intuitive meaning. Experimental results on the YALE B, COIL-100, and MNIST datasets demonstrate the effectiveness of the proposed method.
\end{abstract}

\begin{IEEEkeywords}
Dimensionality reduction, subspace learning, projection matrix, feature extraction.
\end{IEEEkeywords}

%
\IEEEpeerreviewmaketitle

\section{Introduction}
%
%
%
%
\IEEEPARstart{G}{enerally}, input image (concatenated as a  vector) of a computer vision system is high-dimensional. It is known that the curse of the dimensionality occurs when the number of training samples per class is smaller than the dimension of the samples. On the one hand, the high dimension of the data gives arise to the overfitting problem and limits the generalization ability of the system. On the other hand, the high dimension of the data leads to low efficiency in classifying an image. Therefore, dimensionality reduction is a fundamental task of many applications of computer vision and other pattern recognition.

Linear methods of dimensionality reduction are more efficient \cite{15} than the nonlinear counterparts and are basis of the nonlinear methods. Therefore, this paper focuses on linear methods.

The main goal of linear dimensionality reduction method is learning a projection matrix from high-dimensional training data with a proper criterion and some constraints. Low-dimensional representation is achieved by the projection matrix whose number of columns is smaller than the dimension of the input data. To learn the projection matrix, it is required by almost all methods that the relationship of the high-dimensional training samples is known or computed. The relationship information includes which samples are neighbors and the similarity (affinity or connection weight) between a pair of samples. For example, in classical LPP (Locality Preserving Projection) \cite{1}, a predefined number of neighbors are selected according to the Euclidian distance in high-dimensional space and the similarity (affinity) between each pair of the samples are computed using an exponential function. As a supervised algorithm, LFDA (Local Fisher Discriminant Analysis) \cite{3} computes the neighbors and the similarity between them in class-wise manner.

It is note that in classical methods such as LPP LFDA process of selecting neighbors and computing the similarity is independently from the process of learning the projection matrix. We argue that the neighbors in high-dimensional space are not necessarily neighbors in the underlying low-dimensional space and the similarity obtained in the high-dimensional space can not hence capture the intrinsic similarity. A toy example is shown in Fig. 1. In the original high-dimensional (i.e., two-dimensional) space, one feature ${{x}_{1}}$ stands for lightness and the other feature ${{x}_{2}}$ stands for length. 
\begin{figure}[!hbt]
	\centering
	\includegraphics[width=3.5in]{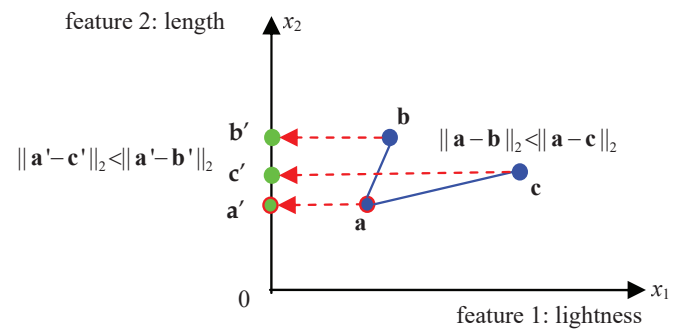}
	\caption{Because  $||\mathbf{a}-\mathbf{b}|{{|}_{2}}<||\mathbf{a}-\mathbf{c}|{{|}_{2}}$, the nearest neighbor of $\mathbf{a}$ in two-dimensional space is $\mathbf{b}$. When projected onto the one-dimensional space (vertical axis),  $\mathbf{a}$, $\mathbf{b}$, and $\mathbf{c}$ are transformed to $\mathbf{a}'$, $\mathbf{b}'$, and $\mathbf{c}'$, respectively. In the one-dimensional space, the nearest neighbor of $\mathbf{a}'$  is $\mathbf{c}'$ instead of  $\mathbf{b}'$ because $||\mathbf{a}'-\mathbf{c}'|{{|}_{2}}<||\mathbf{a}'-\mathbf{b}'|{{|}_{2}}$.}
	\label{fig1}
\end{figure}Assume that the lightness feature is unstable, which is true in many applications. Now compare the nearest neighbors of $\mathbf{a}$ in the original two-dimensional space spanned by axis  ${{x}_{1}}$ and axis ${{x}_{2}}$ and a one-dimensional space where the samples can be correctly classified by the classifier of Nearest Neighbor. Because  $||\mathbf{a}-\mathbf{b}|{{|}_{2}}<||\mathbf{a}-\mathbf{c}|{{|}_{2}}$, the nearest neighbor of $\mathbf{a}$in two-dimensional space is $\mathbf{b}$. Because the feature of lightness is not discriminative, the three samples (i.e., $\mathbf{a}$, $\mathbf{b}$, and $\mathbf{c}$) are transformed to the one-dimensional space spanned the vertical axis.  Specifically, $\mathbf{a}$, $\mathbf{b}$, and $\mathbf{c}$ are transformed to $\mathbf{a}'$, $\mathbf{b}'$, and $\mathbf{c}'$, respectively. In the one-dimensional space, the nearest neighbor of $\mathbf{a}'$ is $\mathbf{c}'$ instead of  $\mathbf{b}'$ because $||\mathbf{a}'-\mathbf{c}'|{{|}_{2}}<||\mathbf{a}'-\mathbf{b}'|{{|}_{2}}$. The toy example demonstrates that the neighbors obtained in high-dimensional space might not be correct and computing the neighbors in proper low-dimensional space might be better for the purpose of classification.

It is inspired by the toy example shown in Fig. 1 that the similarity computed in high-dimensional space can not directly be used as the similarity in the low-dimensional space. That is, the similarity should not be fixed and should vary with the low-dimensional representation. Based on this insight, we propose an objective function where both the similarity and the projection matrix for mapping high-dimensional space to low-dimensional space are unknown variables. In summary, the novelties and contributions of the paper are as follows.

\begin{enumerate}
	\renewcommand{\labelenumii}{(\arabic{enumii})}
	\item We formulate both the similarity of each pair of samples and the projection matrix as variables to be found. In traditional methods, only the projection matrix is expressed as a variable whereas the similarity is fixed and is computed in the original high-dimensional space. By jointing optimizing the similarity and the projection matrix, it is expected that our method is able to yields more optimal solutions.  Therefore, the proposed similarity is classification-oriented whereas existing similarity is feature-oriented.
	\item In our method, the proposed similarity satisfies sum-to-one constraint and non-negative constraint. The sum of the similarities between one sample and all the other samples equals to one. Thus, the proposed non-negative similarity satisfies the properties of the probability. Within each class, this condition makes that each sample can be a neighbor of the other sample. Theoretical analysis shows that the optimal similarity is a function of the projection matrix.
	\item In the proposed unified objective function, there is a regularization parameter for the similarity norm penalty term. The penalty term makes the similarity is sparse to some extent. That is, not all samples are neighbors of one sample and only a fraction of the samples are neighbors of the sample. Instead of empirically setting the regularization parameter, we treat it as a variable to be optimized. Theoretical analysis shows that the regularization parameter is related to the sum of the squared distances of neighbors in low-dimensional space. That is, the optimal regularization parameter is also a function of the projection matrix.
\end{enumerate}

The remainder of the paper is organized as follows: In Section 2, related work is discussed. The proposed SLNP algorithm is described in Section 3. Experimental results are given in Section 4 before summarizing and concluding in Section 5.

 

\section{Related Work}
There are many dimensionality reduction methods \cite{15}. The methods can be divided into supervised, unsupervised \cite{LiuLi_TNNLS2016}, and semi-supervised methods from the point of view of whether or not and how the class labels are utilized. The proposed method belongs to the supervised category. According to how the similarity between samples is obtained and used, the dimensionality reduction methods can be divided into two categories: methods with label-oriented similarity \cite{8,13} and methods with feature-oriented similarity \cite{7,10,11,14}. Because our method differs from existing methods from the point of view of similarity between samples, in this section we mainly review the methods with label-oriented similarity and the methods with feature-oriented similarity.
 
Note that beyond of the scope of this paper there are several classical kinds of dimensionality reduction methods: manifold-based methods \cite{23,24,Li_TNNLS2017,Wang_TNNLS2017}, tensor-based methods \cite{19,20}, probabilistic methods \cite{21,22}, covariance based methods \cite{16,17,18}, non-negative methods \cite{26,27,LiuTongliang_TNNLS2016}, and sparseness and low-rank based methods \cite{28,29,30}.


\subsection{Methods with Label-Oriented Similarity}
In the dimensionality reduction method with label-oriented similarity, the similarity between two samples depends only on their labels. Generally, all pairs of samples share the same similarity. For supervised method, all pairs of samples in each class have the same similarity and the similarity in one class can be either equal to or unequal to the similarity in another class. Representative supervised methods are LDA (Linear Discriminant Analysis) \cite{8} and its variants \cite{25}. For unsupervised method, all pairs of samples in the whole training set have the same similarity. Representative unsupervised methods are PCA (Principal Component Analysis) and its variants \cite{21}.

\textsl{PCA}. Suppose that the training set have $N$ samples: ${{\mathbf{x}}_{1}}$,  ${{\mathbf{x}}_{2}}$, \ldots,  ${{\mathbf{x}}_{N}}$. Let $\mathbf{w}$ be a basis vector (a.k.a., projection vector) and $\mathbf{w}^{*}$ be optimal solution of $\mathbf{w}$ be used for dimensionality reduction. PCA learns the optimal basis vector $\mathbf{w}^{*}$ from the training set based on the least squares reconstruction criterion or equivalently the maximum variance criterion: 
\begin{equation}\begin{split}
\mathbf{w}^{*}&=\arg \underset{{{\mathbf{w}}^{T}}\mathbf{w}=1}{\mathop{\min }}\,{{\mathbf{w}}^{T}}\mathbf{Cw}\\
&=\arg \underset{{{\mathbf{w}}^{T}}\mathbf{w}=1}{\mathop{\min }}\,\frac{1}{N}{{\mathbf{w}}^{T}}\sum\nolimits_{i=1}^{N}{({{\mathbf{x}}_{i}}-\mathbf{\bar{x}}){{({{\mathbf{x}}_{i}}-\mathbf{\bar{x}})}^{T}}}\mathbf{w}\\
&=\arg \underset{{{\mathbf{w}}^{T}}\mathbf{w}=1}{\mathop{\min }}\,\sum\limits_{i\ne j}{\frac{1}{N}{{({{\mathbf{w}}^{T}}{{\mathbf{x}}_{i}}-{{\mathbf{w}}^{T}}{{\mathbf{x}}_{j}})}^{2}}},
\end{split}\end{equation}
where ${\mathbf{C}=(1/N)\sum\nolimits_{i=1}^{N}{({{\mathbf{x}}_{i}}-\mathbf{\bar{x}}){{({{\mathbf{x}}_{i}}-\mathbf{\bar{x}})}^{T}}}}$ is the covariance matrix, ${\mathbf{\bar{x}}=(1/N)\sum\nolimits_{i=1}^{N}{{{\mathbf{x}}_{i}}}}$ is the mean of the $N$ training samples, and ${{{\mathbf{w}}^{T}}\mathbf{w}=||\mathbf{w}||_{2}^{2}=1}$ constrains the norm of the basis vector. Defining 
\begin{equation}{{{s}_{ij}}=\frac{1}{N}},\end{equation}
the problem of PCA can be expressed as
\begin{equation}
\mathbf{w}^{*}=\arg \underset{{{\mathbf{w}}^{T}}\mathbf{w}=1}{\mathop{\min }}\,\sum\limits_{i\ne j}{{{s}_{ij}}{{({{\mathbf{w}}^{T}}{{\mathbf{x}}_{i}}-{{\mathbf{w}}^{T}}{{\mathbf{x}}_{j}})}^{2}}}.
\end{equation}

From the point of view of graph embedding, ${{s}_{ij}}=1/N$ implies that the similarities are equal for any pair of samples of the training set. The label-oriented similarity can also be interpreted that all the samples are neighbors of one sample and there is no difference in similarities.

\textsl{LDA}. Suppose that the $N$ training samples $\{{{\mathbf{x}}_{1}} , {{\mathbf{x}}_{2}},$ $ \ldots,{{\mathbf{x}}_{N}}\}$ are divided into  $C$ different classes and the class labels are $\{1, 2, \ldots, C\}$. The class label of a sample ${{\mathbf{x}}_{i}}$ is denoted by $l({{\mathbf{x}}_{i}})$ with $l({{\mathbf{x}}_{i}})\in \{1, \ldots,C\}$. The number of samples class $i$ is ${{N}_{i}}$. LDA aims at finding the optimal basis vector $\mathbf{w}^{*}$ that maximizes the Rayleigh coefficient or equivalently minimizes the inverse of the Rayleigh coefficient:
\begin{equation}
\mathbf{w}^{*}=\arg \min \frac{{{\mathbf{w}}^{T}}{{\mathbf{S}}_{w}}\mathbf{w}}{{{\mathbf{w}}^{T}}{{\mathbf{S}}_{b}}\mathbf{w}},
\end{equation}
where ${{\mathbf{S}}_{w}}$ and ${{\mathbf{S}}_{b}}$ are the within-class scatter matrix and the between-class scatter matrix:
\begin{align}
&{{\mathbf{S}}_{w}}=\sum\limits_{j=1}^{C}{\sum\limits_{l({{\mathbf{x}}_{i}})=j}{({{\mathbf{x}}_{i}}-{{{\mathbf{\bar{x}}}}_{j}}){{({{\mathbf{x}}_{i}}-{{{\mathbf{\bar{x}}}}_{j}})}^{T}}}},\\
&{{\mathbf{S}}_{b}}=\frac{1}{C}\sum\limits_{j=1}^{C}{({{{\mathbf{\bar{x}}}}_{i}}-\mathbf{\bar{x}}){{({{{\mathbf{\bar{x}}}}_{i}}-\mathbf{\bar{x}})}^{T}}}.
\end{align}
In (5) and (6), $\mathbf{\bar{x}}$ is the mean of all the training samples and ${{\mathbf{\bar{x}}}_{i}}$ is the mean vector of the samples of class $i$:
\begin{equation}
{{\mathbf{\bar{x}}}_{i}}=\sum\limits_{l({{\mathbf{x}}_{k}})=i}{{{\mathbf{x}}_{k}}}.
\end{equation}
Substituting (7) into (5) and (6) yields
\begin{flalign}
&{\medmuskip -2mu\thickmuskip -2mu\thinmuskip -2mu{\mathbf{S}}_{w}= \begin{cases}
\dfrac{1}{2}\sum\limits_{i,j=1}^{N}{\dfrac{1}{{{n}_{l({{x}_{i}})}}}({{\mathbf{x}}_{i}}-{{\mathbf{x}}_{j}}){{({{\mathbf{x}}_{i}}-{{\mathbf{x}}_{j}})}^{T}},} & \text{if }l({{\mathbf{x}}_{j}})=l({{\mathbf{x}}_{i}})  \\
0, & \text{if }l({{\mathbf{x}}_{j}})\ne l({{\mathbf{x}}_{i}})  \\
\end{cases},}\\
\intertext{and}
&{\medmuskip -2mu\thickmuskip -2mu\thinmuskip -2mu{{\mathbf{S}}_{b}}= \begin{cases}
\dfrac{1}{2}\sum\limits_{i,j=1}^{N}{\left( \frac{1}{N}-\frac{1}{{{N}_{l({{x}_{i}})}}} \right)({{\mathbf{x}}_{i}}-{{\mathbf{x}}_{j}}){{({{\mathbf{x}}_{i}}-{{\mathbf{x}}_{j}})}^{T}},} & \text{if }l({{\mathbf{x}}_{j}})=l({{\mathbf{x}}_{i}})  \\
\dfrac{1}{2}\sum\limits_{i,j=1}^{N}{\frac{1}{N}({{\mathbf{x}}_{i}}-{{\mathbf{x}}_{j}}){{({{\mathbf{x}}_{i}}-{{\mathbf{x}}_{j}})}^{T}},} & \text{if }l({{\mathbf{x}}_{j}})\ne l({{\mathbf{x}}_{i}})  \\
\end{cases}\mspace{-5mu}, }
\end{flalign}
respectively. 

Defining respectively the similarity $s_{ij}^{w}$ for within-class scatter and the similarity  $s_{ij}^{w}$ as 
\begin{flalign}
&{s_{ij}^{w}=\begin{cases}
\dfrac{1}{{{N}_{l({{x}_{i}})}}} & \text{if }l({{\mathbf{x}}_{j}})=l({{\mathbf{x}}_{i}})  \\
0 & \text{if }l({{\mathbf{x}}_{j}})\ne l({{\mathbf{x}}_{i}})  \\
\end{cases},} \\
\intertext{and}
&{s_{ij}^{b}= \begin{cases}
\dfrac{1}{N}-\dfrac{1}{{{N}_{l({{x}_{i}})}}} & \text{if }l({{\mathbf{x}}_{j}})=l({{\mathbf{x}}_{i}})  \\
\dfrac{1}{N} & \text{if }l({{\mathbf{x}}_{j}})\ne l({{\mathbf{x}}_{i}})  \\
\end{cases}. }
\end{flalign}

With the similarities ${s_{ij}^{w}}$ and ${s_{ij}^{b}}$, LDA can be expressed as the following optimization problem:

\begin{equation}\begin{split}
\mathbf{w}^{*}&=\arg \min \frac{{{\mathbf{w}}^{T}}{{\mathbf{S}}_{w}}\mathbf{w}}{{{\mathbf{w}}^{T}}{{\mathbf{S}}_{b}}\mathbf{w}} \\ 
&=\arg \min \frac{\sum\limits_{i\ne j}^{N}{s_{ij}^{w}{{({{\mathbf{w}}^{T}}{{\mathbf{x}}_{i}}-{{\mathbf{w}}^{T}}{{\mathbf{x}}_{j}})}^{2}}}}{\sum\limits_{i\ne j}^{N}{s_{ij}^{b}{{({{\mathbf{w}}^{T}}{{\mathbf{x}}_{i}}-{{\mathbf{w}}^{T}}{{\mathbf{x}}_{j}})}^{2}}}}. \\ 
\end{split}\end{equation}
The similarities $s_{ij}^{w}$ and $s_{ij}^{b}$ in (12) are related to the class labels. 

\subsection{Methods with Feature-Oriented Similarity}
The label-oriented similarity of two samples is completely determined by the labels of the samples. Therefore, the label-oriented similarity is irrelevant to the features of the samples. However, the values of the feature vectors are important for measuring the similarity of two samples. Generally speaking, feature-oriented similarity is superior to label-oriented similarity because not only class labels (if given) but also features are used for computing similarity. Representative feature-oriented methods include LPP (a.k.a., Laplacianface in the community of face recognition) \cite{1}, MFA (Marginal Fisher Analysis) \cite{2}, and LFDA (Local Fisher Discriminant Analysis) \cite{3}, SOLDE (Stable Orthogonal Local Discriminant Embedding) \cite{14}, JGLDA (Joint Global and Local Structure Discriminant Analysis) \cite{7}.

\textsl{LPP}. In LPP, the similarity ${{s}_{ij}}$ between ${{\mathbf{x}}_{i}}$ and ${{\mathbf{x}}_{j}}$ is :
\begin{equation}
{{s}_{ij}}=\exp \left( -\frac{{{({{\mathbf{x}}_{i}}-{{\mathbf{x}}_{j}})}^{2}}}{t} \right).
\end{equation}

It can be seen from (13) that the similarity is a function of the difference between the feature vector ${{\mathbf{x}}_{i}}$ and feature vector ${{\mathbf{x}}_{j}}$. Therefore, the similarity in LPP is called feature-oriented. The similarity also depends on the parameter $t$ which is usually empirically chosen. 

With the feature-oriented similarity, the optimal projection vector $\mathbf{w}^{*}$ is obtained by solving the following optimization problem:

\begin{equation}
\medmuskip -2mu\thickmuskip -2mu\thinmuskip -2mu
\mathbf{w}^{*}=\arg {\underset{\sum\limits_{i=1}^{N}{\left( \sum\limits_{j=1}^{N}{{{s}_{ij}}} \right){{({{\mathbf{w}}^{T}}{{\mathbf{x}}_{i}})}^{2}}=1}}{\mathop{\min }}}\sum\limits_{i\ne j}^{N}{{{s}_{ij}}{{({{\mathbf{w}}^{T}}{{\mathbf{x}}_{i}}-{{\mathbf{w}}^{T}}{{\mathbf{x}}_{j}})}^{2}}}.
\end{equation}

The effect of weighting the difference ${{({{\mathbf{w}}^{T}}{{\mathbf{x}}_{i}}-{{\mathbf{w}}^{T}}{{\mathbf{x}}_{j}})}^{2}}$ with the feature-oriented similarity  is to ensure that, if ${{\mathbf{x}}_{i}}$ and ${{\mathbf{x}}_{j}}$ are close in the original high-dimensional space, then their low-dimensional representations ${{\mathbf{w}}^{T}}{{\mathbf{x}}_{i}}$ and ${{\mathbf{w}}^{T}}{{\mathbf{x}}_{j}}$ are close as well \cite{1}.

\textsl{LFDA}. The similarity in LFDA can be seen as a combination of the label-oriented similarity of LDA (i.e., Eq. (10) and Eq. (11)) and the feature-oriented similarity of LPP (i.e., Eq. (13)). Specifically, the similarity $s_{ij}^{w}$ for the within-class scatter and the similarity $s_{ij}^{b}$ for the between-class scatter are:
\begin{flalign}
&{s_{ij}^{w}= \begin{cases}
\dfrac{{{s}_{ij}}}{{{N}_{l({{x}_{i}})}}} & \text{if }l({{\mathbf{x}}_{j}})=l({{\mathbf{x}}_{i}})  \\
0 & \text{if }l({{\mathbf{x}}_{j}})\ne l({{\mathbf{x}}_{i}})  \\
\end{cases},}
\intertext{and}
&{s_{ij}^{b}=\begin{cases}
{{s}_{ij}}\left( \dfrac{1}{N}-\dfrac{1}{{{N}_{l({{x}_{i}})}}} \right) & \text{if }l({{\mathbf{x}}_{j}})=l({{\mathbf{x}}_{i}})  \\
\dfrac{1}{N} & \text{if }l({{\mathbf{x}}_{j}})\ne l({{\mathbf{x}}_{i}})  \\
\end{cases}.}
\end{flalign}
respectively. The ${{s}_{ij}}$ in (15) and (16) is the same as the ${{s}_{ij}}$ in (11). Because the feature-oriented similarity in a local way, the resulting similarity of LFDA makes LFDA is capable of dealing with multimodal class which is composed of samples of several separate clusters. With the similarities expressed in (15) and (16), the optimization problem of LFDA is in the same form of (12).

Investigating the formulas (3), (12), and (14), one can see that the computation of the label-oriented and feature-oriented similarities is prior to the computation of the projection vectors.

\section{Proposed Method}

The analysis in Section 2 shows that existing methods employ fixed similarities for learning projection vectors (matrix). The computation of the similarities is prior to and independent to the computation of the projection vectors. As shown in Fig. 2, the traditional label-oriented similarity and feature-oriented similarity can be categorized as fixed similarity. In this paper, we propose variable similarity for learning better projection vectors. The proposed variable similarity varies with projection vector and is classification-oriented. Both the variable similarity and projection vector are formulated in a unified objective function with proper constraints on the similarity and projection vector. 

\begin{figure}[!htb]
	\centering
	\includegraphics[width=3.5in]{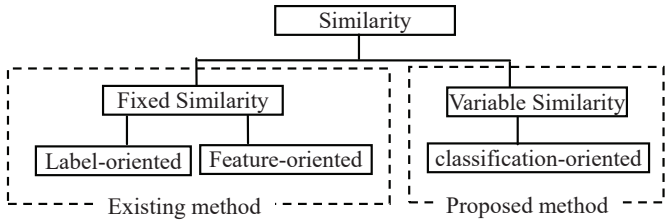}
	\caption{Fixed similarity versus the proposed variable similarity.}
	\label{fig2}
\end{figure}

In this section, we begin by formulating the objective function and the constraints of the proposed method followed by describing how to solve the corresponding optimization problem. 

\subsection{Objective Function and Constraints with Variable Similarity}

\subsubsection{Data}

The training stage is to learn an optimal projection matrix $\mathbf{W}\in {{\mathbb{R}}^{D\times d}}$ from the $N$ training samples  $\mathbf{X}=\{{{\mathbf{x}}_{11}}, {{\mathbf{x}}_{12}}, \ldots, {{\mathbf{x}}_{1{{N}_{1}}}}, {{\mathbf{x}}_{21}}, {{\mathbf{x}}_{22}}, \ldots,{{\mathbf{x}}_{2{{N}_{2}}}}, \ldots, {{\mathbf{x}}_{C{{N}_{C}}}}\}$ with ${{\mathbf{x}}_{ij}}\in {{\mathbb{R}}^{D\times 1}}$,  $d<D$, and $N=\sum\nolimits_{i=1}^{C}{{{N}_{i}}}$. The $N$ training samples can be divided into $C$ different classes and each class $i$ consists of ${{N}_{i}}$ samples. The subscripts $i$ and $j$ of ${{\mathbf{x}}_{ij}}$ index the class and the sample  in the class, respectively. For the sake of notation simplicity,  it is assumed that ${{N}_{i}}={{N}_{j}}=M$, $i\ne j$. Note that the proposed theory and algorithm work also for ${{N}_{i}}\ne {{N}_{j}}$. The $d$-dimensional representation ${{\mathbf{y}}_{ij}}\in {{\mathbb{R}}^{d}}$ of the $D$-dimensional sample  ${{\mathbf{x}}_{ij}}\in {{\mathbb{R}}^{D}}$ is obtained by ${{\mathbf{y}}_{ij}}={{\mathbf{W}}^{T}}{{\mathbf{x}}_{ij}}$. 

\subsubsection{Similarity}

Let similarity ${{s}_{ijk}}$ denote the similarity between the sample ${{\mathbf{x}}_{ij}}$ and the sample ${{\mathbf{x}}_{ik}}$ of the class $i$. The similarities for class $i$ form a symmetric similarity matrix ${{\mathbf{S}}_{i}}\in {{\mathbb{R}}^{M\times M}}$. The $j$-th column vector ${{\mathbf{s}}_{ij}}\in {{\mathbb{R}}^{M\times 1}}$ stands for the similarities for the $j$-th sample of class $i$ and the $k$-th element of ${{\mathbf{s}}_{ij}}$ is ${{s}_{ijk}}$. The similarity matrices $({{\mathbf{S}}_{1}},{{\mathbf{S}}_{2}}, \ldots, {{\mathbf{S}}_{C}})$ for all the $C$ classes form a similarity tensor $\mathbf{S}\in {{\mathbb{R}}^{C\times N\times N}}$. Traditional methods pre-defined (pre-computed) the similarity according to the class labels or the values (features) of the samples ${{\mathbf{x}}_{ij}}$ and ${{\mathbf{x}}_{ik}}$. In our method, the similarity ${{s}_{ijk}}$ is a variable and satisfies the properties of probability: 
\begin{align}
&0\le {{s}_{ijk}}\le 1,\\
&\sum\limits_{k=1}^{{{N}_{i}}}{{{s}_{ijk}}}=1.
\end{align}

\subsubsection{Objective Function, Constraints and Regularization Term}

The similarity tensor $\mathbf{S}$ (whose elements are ${{s}_{ijk}}$) and the projection matrix $\mathbf{W}$ are obtained by minimizing a unified objective function $J(\mathbf{S},\mathbf{W},\mathbf{R})$:
\begin{equation}
\medmuskip 1mu\thickmuskip 1mu\thinmuskip 1mu
J(\mathbf{S},\mathbf{W},\mathbf{R})\text{=}\sum\limits_{i=1}^{C}{\sum\limits_{j=1}^{{{N}_{i}}}{\sum\limits_{k=1}^{{{N}_{i}}}{({{s}_{ijk}}||{{\mathbf{W}}^{T}}{{\mathbf{x}}_{ij}}-{{\mathbf{W}}^{T}}{{\mathbf{x}}_{ik}}||_{2}^{2}+{{\gamma }_{ij}}s_{ijk}^{2})}}} 
\end{equation}
with non-negative constraints on ${{s}_{ijk}}$
\begin{equation}
{{s}_{ijk}}\ge 0, i=1,\ldots,C, i,j=1,\ldots,{{N}_{i}},
\end{equation}
sum-to-one constraints on ${{s}_{ijk}}$
\begin{equation}
\sum\limits_{k=1}^{{{N}_{i}}}{{{s}_{ijk}}}=1, i=1,\ldots,C,
\end{equation}
and whitening constraints on $\mathbf{W}$
\begin{equation}
\mathbf{W}^{T}{{\mathbf{S}}_{t}}\mathbf{W}=\mathbf{I}.	
\end{equation}   

The non-negative constraints (20) and the sum-to-one constraints (21) guarantee that the similarity ${{s}_{ijk}}$ is a probability. The effect of the whitening constraints on $\mathbf{W}$ is letting the features of the total training samples having the equal variance. In (22), ${{\mathbf{S}}_{t}}$ is the total scatter matrix:
\begin{equation}
{{\mathbf{S}}_{t}}=\sum\limits_{i=1}^{C}{\sum\limits_{j=1}^{{{N}_{i}}}{({{\mathbf{x}}_{ij}}-\mathbf{\bar{x}}){{({{\mathbf{x}}_{ij}}-\mathbf{\bar{x}})}^{T}}}},
\end{equation}	
with $\mathbf{\bar{x}}$ being the mean of the total training samples:
\begin{equation}
\mathbf{\bar{x}}=\frac{1}{N}\sum\limits_{i=1}^{C}{\sum\limits_{j=1}^{{{N}_{i}}}{{{\mathbf{x}}_{ij}}}}.	
\end{equation}

The regularization term (penalty term) ${{\gamma }_{ij}}s_{ijk}^{2}$ is very important for solving meaningful similarities. In some degree, the effect of the regularization term is to let the similarities are sparse. This effect is in line with the intuition that only a small number samples are very similar to one sample and the neighboring samples are in a small region of the sample. The regularization parameters ${{\gamma }_{ij}}$ with $i=1,2,\ldots,C$ and $j=1,2,\ldots,{{N}_{i}}$ form a matrix $\mathbf{R}\in {{\mathbb{R}}^{C\times {{N}_{i}}}}$ with its $ij$ entry being ${{\gamma }_{ij}}$. The $i$-th row of $\mathbf{R}$ are the regularization parameters corresponding to class $i$. We denote the transpose of the $i$-th row of $\mathbf{R}$ by the column vector ${{\mathbf{r}}_{i}}\in {{\mathbb{R}}^{{{N}_{i}}\times 1}}$. We call $\mathbf{R}$ regularization matrix.

Note that the regularization parameter ${{\gamma }_{ij}}$ is also a variable and hence we express the objective function $J(\mathbf{S},\mathbf{W},\mathbf{R})$ as a function of $\mathbf{S}$,  $\mathbf{W}$, and  $\mathbf{R}$. 

\subsection{Optimization}

For the sake of clarity, the optimization problem corre-sponding the objective function (19) and the constraints (20), (21), and (22) is written as
\begin{equation}\begin{split}
&\underset{\mathbf{S},\mathbf{W},\mathbf{R}}{\mathop{\min }}\,\sum\limits_{i=1}^{C}{\sum\limits_{j=1}^{{{N}_{i}}}{\sum\limits_{k=1}^{{{N}_{i}}}{({{s}_{ijk}}||{{\mathbf{W}}^{T}}{{\mathbf{x}}_{ij}}-{{\mathbf{W}}^{T}}{{\mathbf{x}}_{ik}}||_{2}^{2}+{{\gamma }_{ij}}s_{ijk}^{2})}}}.\\
&s.t.\quad{{s}_{ijk}}\ge 0,\text{ }\sum\limits_{k=1}^{{{N}_{i}}}{{{s}_{ijk}}}=1,\text{ }{{\mathbf{W}}^{T}}{{\mathbf{S}}_{t}}\mathbf{W}=\mathbf{I}
\end{split}\end{equation}

The task is to find the optimal similarity tensor $\mathbf{S}\in {{\mathbb{R}}^{C\times N\times N}}$, the projection matrix $\mathbf{W}\in {{\mathbb{R}}^{D\times d}}$, and the regularization parameter $\gamma \in \mathbb{R}$. We propose an alternative algorithm to seek the optimal variables $\mathbf{S}$, $\mathbf{W}$, and $\gamma $ in turn. 

\subsubsection{$\mathbf{S}$-step (Compute $\mathbf{S}$ when $\mathbf{W}$ and ${{\gamma }_{ij}}$ are fixed)}

The goal of  $\mathbf{S}$-step is to learn optimal $\mathbf{S}$-step when  $\mathbf{W}$ and ${{\gamma }_{ij}}$ are fixed. With fixed $\mathbf{W}$ and ${{\gamma }_{ij}}$ the optimization problem is reduced to
\begin{equation}
\medmuskip -1mu\thickmuskip -1mu\thinmuskip -1mu
\underset{\begin{smallmatrix} 
	\mathbf{S} \\ 
	{{s}_{ijk}}\ge 0,\text{ }\sum\limits_{k=1}^{{{N}_{i}}}{{{s}_{ijk}}}=1 
	\end{smallmatrix}}{\mathop{\min }}\,\sum\limits_{i=1}^{C}{\sum\limits_{j=1}^{{{N}_{i}}}{\sum\limits_{k=1}^{{{N}_{i}}}{({{s}_{ijk}}||{{\mathbf{W}}^{T}}{{\mathbf{x}}_{ij}}-{{\mathbf{W}}^{T}}{{\mathbf{x}}_{ik}}||_{2}^{2}+{{\gamma }_{ij}}s_{ijk}^{2})}}}. 
\end{equation}

The three-order tensor $\mathbf{S}$ contains $C$ similarity matrices ${{\mathbf{S}}_{1}}$, ${{\mathbf{S}}_{2}}$, \ldots, ${{\mathbf{S}}_{C}}$. The similarity matrix ${{\mathbf{S}}_{i}}$ consists of the similarities for class $i$ and its $j-k$ entry is  ${{s}_{ijk}}$. Because the similarity matrix  ${{\mathbf{S}}_{i}}$ is independent to other similarity matrices ${{\mathbf{S}}_{j}}$, $j\ne i$, the optimal matrix ${{\mathbf{S}}_{i}}$ can be individually calculated. The optimization problem for ${{\mathbf{S}}_{i}}$ becomes: 
\begin{equation}
\medmuskip -0mu\thickmuskip -0mu\thinmuskip -0mu
\underset{\begin{smallmatrix} 
	{{\mathbf{S}}_{i}} \\ 
	{{s}_{ijk}}\ge 0,\text{ }\sum\limits_{k=1}^{{{N}_{i}}}{{{s}_{ijk}}}=1 
	\end{smallmatrix}}{\mathop{\min }}\,\sum\limits_{j=1}^{{{N}_{i}}}{\sum\limits_{k=1}^{{{N}_{i}}}{({{s}_{ijk}}||{{\mathbf{W}}^{T}}{{\mathbf{x}}_{ij}}-{{\mathbf{W}}^{T}}{{\mathbf{x}}_{ik}}||_{2}^{2}+{{\gamma }_{ij}}s_{ijk}^{2})}}.
\end{equation}

Because the projection matrix $\mathbf{W}$ is fixed and the samples ${{\mathbf{x}}_{ij}}$ and ${{\mathbf{x}}_{ik}}$ are given, the squared distance in $||{{\mathbf{W}}^{T}}{{\mathbf{x}}_{ij}}-{{\mathbf{W}}^{T}}{{\mathbf{x}}_{ik}}||_{2}^{2}$ in the low-dimensional space is a constant which we denote by ${{d}_{ijk}}$:
\begin{equation}
{{d}_{ijk}}\triangleq ||{{\mathbf{W}}^{T}}{{\mathbf{x}}_{ij}}-{{\mathbf{W}}^{T}}{{\mathbf{x}}_{ik}}||_{2}^{2}.
\end{equation}

Then (27) can be written as
\begin{equation}
\medmuskip -1mu\thickmuskip -1mu\thinmuskip -1mu
\begin{split}
\mathbf{S}_{i}^{*}&=\arg \underset{\begin{smallmatrix} 
	{{\mathbf{S}}_{i}} \\
	{{s}_{ijk}}\ge 0,\text{ }\sum\limits_{k=1}^{{{N}_{i}}}{{{s}_{ijk}}}=1 
	\end{smallmatrix}}{\mathop{\min }}\,\sum\limits_{j=1}^{{{N}_{i}}}{\sum\limits_{k=1}^{{{N}_{i}}}{({{d}_{ijk}}{{s}_{ijk}}+{{\gamma }_{ij}}s_{ijk}^{2})}}\\ 
&=\arg \underset{\begin{smallmatrix} 
	{{\mathbf{S}}_{i}} \\
	{{s}_{ijk}}\ge 0,\text{ }\sum\limits_{k=1}^{{{N}_{i}}}{{{s}_{ijk}}}=1 
	\end{smallmatrix}}{\mathop{\min }}\,\sum\limits_{j=1}^{{{N}_{i}}}{\sum\limits_{k=1}^{{{N}_{i}}}{\left[ {{\gamma }_{ij}}{{\left( {{s}_{ijk}}+\frac{1}{2{{\gamma }_{ij}}}{{d}_{ijk}} \right)}^{2}}-\frac{d_{ijk}^{2}}{4{{\gamma }_{ij}}} \right]}}\\
&=\arg \underset{\begin{smallmatrix} 
	{{\mathbf{S}}_{i}} \\
	{{s}_{ijk}}\ge 0,\text{ }\sum\limits_{k=1}^{{{N}_{i}}}{{{s}_{ijk}}}=1 
	\end{smallmatrix}}{\mathop{\min }}\,\sum\limits_{j=1}^{{{N}_{i}}}{\sum\limits_{k=1}^{{{N}_{i}}}{\left[ {{\gamma }_{ij}}{{\left( {{s}_{ijk}}+\frac{1}{2{{\gamma }_{ij}}}{{d}_{ijk}} \right)}^{2}} \right]}}\\
&=\arg \underset{\begin{smallmatrix} 
	{{\mathbf{S}}_{i}} \\
	{{s}_{ijk}}\ge 0,\text{ }\sum\limits_{k=1}^{{{N}_{i}}}{{{s}_{ijk}}}=1 
	\end{smallmatrix}}{\mathop{\min }}\,\sum\limits_{j=1}^{{{N}_{i}}}{{{\gamma }_{ij}}\sum\limits_{k=1}^{{{N}_{i}}}{{{\left( {{s}_{ijk}}+\frac{1}{2{{\gamma }_{ij}}}{{d}_{ijk}} \right)}^{2}}}}.
\end{split}\end{equation}
Define 
\begin{align}
&{{q}_{ijk}}\triangleq \frac{1}{2{{\gamma }_{ij}}}{{d}_{ijk}},\\
&{{\mathbf{q}}_{ij}}\triangleq \!\![\!\!\text{ }{{q}_{ij1}},{{q}_{ij2}},\ldots ,{{q}_{ijM}}{{]}^{T}},\\
&{{\mathbf{d}}_{ij}}\triangleq \!\![\!\!\text{ }{{d}_{ij1}},{{d}_{ij2}},\ldots ,{{d}_{ijM}}{{]}^{T}},
\end{align}
then the last line of (29) can be written as a minimization problem of quadratic function:
\begin{equation}
\mathbf{S}_{i}^{*}=\arg \underset{\begin{smallmatrix} 
	{{\mathbf{S}}_{i}} \\ 
	{{s}_{ijk}}\ge 0,\text{ }\sum\limits_{k=1}^{{{N}_{i}}}{{{s}_{ijk}}}=1 
	\end{smallmatrix}}{\mathop{\min }}\,\sum\limits_{j=1}^{{{N}_{i}}}{{{\gamma }_{ij}}\sum\limits_{k=1}^{{{N}_{i}}}{{{\left( {{s}_{ijk}}+{{q}_{ijk}} \right)}^{2}}}}.
\end{equation}

Because the similarity vector  ${{\mathbf{s}}_{ij}}$ is not related to the similarity vector ${{\mathbf{s}}_{ik}}$ for $j\ne k$, each similarity vector can be computed separately: 
\begin{equation}\begin{split}
\mathbf{s}_{ij}^{*}&=\arg \underset{\begin{smallmatrix} 
	{{\mathbf{s}}_{ii}} \\ 
	{{s}_{ijk}}\ge 0,\text{ }\sum\limits_{k=1}^{{{N}_{i}}}{{{s}_{ijk}}}=1 
	\end{smallmatrix}}{\mathop{\min }}\,\sum\limits_{k=1}^{{{N}_{i}}}{{{\left( {{s}_{ijk}}+{{q}_{ijk}} \right)}^{2}}} \\                                 
&=\arg \underset{\begin{smallmatrix} 
	{{\mathbf{s}}_{ii}} \\ 
	{{s}_{ijk}}\ge 0,\text{ }\sum\limits_{k=1}^{{{N}_{i}}}{{{s}_{ijk}}}=1 
	\end{smallmatrix}}{\mathop{\min }}\,||{{\mathbf{s}}_{ij}}+{{\mathbf{q}}_{ij}}||_{2}^{2}.
\end{split}\end{equation}

Because $||{{\mathbf{s}}_{ij}}+{{\mathbf{q}}_{ij}}||_{2}^{2}$ is a convex function (quadratic function), the inequality constraints ${{s}_{ijk}}\ge 0$ is also convex, and the equality constraint  $\sum\limits_{k=1}^{{{N}_{i}}}{{{s}_{ijk}}}=1$ is an affinity function, one can adopt the technique of Lagrangian multiplier to convert to the constrained optimization problem to the unconstrained optimization problem whose objective function $L({{\mathbf{s}}_{ij}},\eta ,\mathbf{b})$  is: 
\begin{equation}
L({{\mathbf{s}}_{ij}},\eta ,\mathbf{b})=\frac{1}{2}||{{\mathbf{s}}_{ij}}+{{\mathbf{q}}_{ij}}||_{2}^{2}-\eta (\mathbf{s}_{ij}^{T}\mathbf{1}-1)-{{\mathbf{b}}^{T}}{{\mathbf{s}}_{ij}}.
\end{equation} 
In (35), $\eta \ge 0$ and $\mathbf{b}\ge 0$ are the Largrangian multipliers, $\mathbf{1}$ is the vector with each element being 1 and its dimension identical to that of ${{\mathbf{s}}_{ij}}$. The Karush-Kuhn-Tucker (KKT) condition 
\begin{equation}
\nabla \frac{1}{2}||{{\mathbf{s}}_{ij}}+{{\mathbf{q}}_{ij}}||_{2}^{2}-\eta \nabla (\mathbf{s}_{ij}^{T}\mathbf{1}-1)-\nabla {{\mathbf{b}}^{T}}{{\mathbf{s}}_{ij}}=0
\end{equation}
for optimizing (36) results in a feasible minimizer 
\begin{equation}
\mathbf{s}_{ij}^{*}=-{{\mathbf{q}}_{ij}}+\eta +\mathbf{b}=-\frac{{{\mathbf{d}}_{ij}}}{2{{\gamma }_{ij}}}+\eta +\mathbf{b}
\end{equation}
with the constraints $\mathbf{s}_{ij}^{*}\ge 0$, $\mathbf{s}_{ij}^{T}\mathbf{1}=1$, and $\mathbf{b}\ge 0$. For the sake of simplicity, we let $\mathbf{b}=0$. The corresponding feasible minimizer becomes 
\begin{equation}
\mathbf{s}_{ij}^{*}=-{{\mathbf{q}}_{ij}}+\eta =-\frac{{{\mathbf{d}}_{ij}}}{2{{\gamma }_{ij}}}+\eta
\end{equation}
with the constraint  being $\mathbf{s}_{ij}^{*}\ge 0$ and $\mathbf{s}_{ij}^{T}\mathbf{1}=1$.

\subsubsection{$\gamma $-step (Compute $\gamma $ when $\mathbf{W}$ is fixed)}

In (38), there are two unknown parameters: $\eta $ and ${{\gamma }_{ij}}$. Now the question is how to determine $\eta $ and ${{\gamma }_{ij}}$ under the constraints $\mathbf{s}_{ij}^{*}\ge 0$ and $\mathbf{s}_{ij}^{T}\mathbf{1}=1$.  Because $\mathbf{s}_{ij}^{*}$ is a function of $\eta $ and ${{\gamma }_{ij}}$, the optimization problem is transformed from (27) to:
\begin{equation}
\medmuskip -1mu\thickmuskip -1mu\thinmuskip -1mu
\underset{\begin{smallmatrix} 
	{{\gamma }_{ij}},\eta  \\ 
	{{s}_{ijk}}\ge 0,\text{ }\sum\limits_{k=1}^{{{N}_{i}}}{{{s}_{ijk}}}=1 
	\end{smallmatrix}}{\mathop{\min }}\,\sum\limits_{i=1}^{C}{\sum\limits_{j=1}^{{{N}_{i}}}{\sum\limits_{k=1}^{{{N}_{i}}}{({{s}_{ijk}}||{{\mathbf{W}}^{T}}{{\mathbf{x}}_{ij}}-{{\mathbf{W}}^{T}}{{\mathbf{x}}_{ik}}||_{2}^{2}+{{\gamma }_{ij}}s_{ijk}^{2})}}}.
\end{equation}
We first state how to compute the optimal value of $\eta $. Then the low bound and high bound of ${{\gamma }_{ij}}$ are derived. Finally, the method of calculating the optimal ${{\gamma }_{ij}}$ within the bounds is described. 

\textbf{Computation of Optimal $\eta $}. Because of the sum-to-one constraint
\begin{equation}
\sum\limits_{k=1}^{M}{{{s}_{ijk}}}=\sum\limits_{k=1}^{K}{{{s}_{ijk}}}=1,	
\end{equation}
it holds that
\begin{equation}
\sum\limits_{k=1}^{K}{{{s}_{ijk}}}=\sum\limits_{k=1}^{K}{\left( -\frac{{{d}_{ijk}}}{2{{\gamma }_{ij}}}+\eta  \right)}=1. 	
\end{equation}
Therefore, the parameter $\eta $ can be determined by
\begin{equation}
\eta =\frac{1}{K}\left( \frac{1}{2{{\gamma }_{ij}}}\sum\limits_{k=1}^{K}{{{d}_{ijk}}}+1 \right).
\end{equation}
Eq. (42) shows that $\eta $ is also a function of ${{\gamma }_{ij}}$.

\textbf{Computation of Low and High Bounds of ${{\gamma }_{ij}}$}. In order to guarantee $\mathbf{s}_{ij}^{*}\ge 0$, it is reasonably assumed that  the similarity ${{s}_{ijk}}>0$ for the low-dimensional samples ${{\mathbf{W}}^{T}}{{\mathbf{x}}_{ik}}$ which are the $K$ nearest neighbors of the low-dimensional sample ${{\mathbf{W}}^{T}}{{\mathbf{x}}_{ij}}$. The distance ${{d}_{ijk}}=||{{\mathbf{W}}^{T}}{{\mathbf{x}}_{ij}}-{{\mathbf{W}}^{T}}{{\mathbf{x}}_{ik}}|{{|}_{2}}$ is used for determining neighbors of ${{\mathbf{W}}^{T}}{{\mathbf{x}}_{ij}}$. Without loss of generality, assume that the distances are in ascent order (i.e., ${{d}_{ij1}}\le {{d}_{ij2}}\le \cdots \le {{d}_{ijK}}\le {{d}_{ij(K+1)}}\le \cdots \le {{d}_{ijM}}$). Consequently, we have ${{s}_{ijk}}>0$ for $k=1, \ldots ,K$ and ${{s}_{ijk}}=0$ for $k=K+1,K+2, \ldots ,M$:
\begin{flalign}\begin{cases}
{{s}_{ijk}}=-\dfrac{{{d}_{ijk}}}{2{{\gamma }_{ij}}}+\eta >0 & k\le K,  \\
{{s}_{ijk}}=-\dfrac{{{d}_{ijk}}}{2{{\gamma }_{ij}}}+\eta =0 & k>K.  \\
\end{cases}\end{flalign}

Now the only unknown parameter is ${{\gamma }_{ij}}$. Substituting (42) into (43) yields
\begin{flalign}\begin{cases}
{{\gamma }_{ij}}>\dfrac{K}{2}{{d}_{ijk}}-\dfrac{1}{2}\sum\limits_{k=1}^{K}{{{d}_{ijk}}} & k\le K,  \\
{{\gamma }_{ij}}<\dfrac{K}{2}{{d}_{ij(k+1)}}-\dfrac{1}{2}\sum\limits_{k=1}^{K}{{{d}_{ijk}}} & k>K.  \\
\end{cases}\end{flalign}

The inequalities (43) can be reduced to
\begin{equation}
\dfrac{K}{2}{{d}_{ijk}}-\dfrac{1}{2}\sum\limits_{k=1}^{K}{{{d}_{ijk}}\le }{{\gamma }_{ij}}\le \dfrac{K}{2}{{d}_{ij(k+1)}}-\dfrac{1}{2}\sum\limits_{k=1}^{K}{{{d}_{ijk}}}.
\end{equation}

Inequality (45) gives a low bound and a high bound for selecting  ${{\gamma }_{ij}}$. Note that both the low bound and the high bound are non-negative.

\textbf{Computation of the Optimal ${{\gamma }_{ij}}$ within the Bound}. Now we describe how to obtain the optimal ${{\gamma }_{ij}}$ within the bounds given in (45).

Because ${{s}_{ijk}}>0$ for $k=1,\ldots ,K$ and ${{s}_{ijk}}=0$ for $k=K+1,K+2,\ldots ,M$, the objective function of (26) can be written as:
\begin{equation}\begin{split}
&\sum\limits_{i=1}^{C}{\sum\limits_{j=1}^{{{N}_{i}}}{\sum\limits_{k=1}^{{{N}_{i}}}{({{s}_{ijk}}||{{\mathbf{W}}^{T}}{{\mathbf{x}}_{ij}}-{{\mathbf{W}}^{T}}{{\mathbf{x}}_{ik}}||_{2}^{2}+{{\gamma }_{ij}}s_{ijk}^{2})}}} \\ 
&=\sum\limits_{i=1}^{C}{\sum\limits_{j=1}^{{{N}_{i}}}{\sum\limits_{k=1}^{{{N}_{i}}}{({{s}_{ijk}}{{d}_{i}}_{jk}+{{\gamma }_{ij}}s_{ijk}^{2})}}} \\ 
&=\sum\limits_{i=1}^{C}{\sum\limits_{j=1}^{{{N}_{i}}}{\sum\limits_{k=1}^{K}{({{s}_{ijk}}{{d}_{i}}_{jk}+{{\gamma }_{ij}}s_{ijk}^{2})}}}. 
\end{split}\end{equation}

Substituting (38) into the last line of (46) yields:
\begin{equation}\begin{split}
& \sum\limits_{i=1}^{C}{\sum\limits_{j=1}^{{{N}_{i}}}{\sum\limits_{k=1}^{K}{({{s}_{ijk}}{{d}_{i}}_{jk}+{{\gamma }_{ij}}s_{ijk}^{2})}}} \\ 
& =\sum\limits_{i=1}^{C}{\sum\limits_{j=1}^{{{N}_{i}}}{\sum\limits_{k=1}^{K}{\left[ \left( -\frac{{{d}_{jk}}}{2{{\gamma }_{ij}}}+\eta  \right){{d}_{ijk}}+{{\gamma }_{ij}}{{\left( -\frac{{{d}_{ijk}}}{2{{\gamma }_{ij}}}+\eta  \right)}^{2}} \right]}}} \\ 
& =\sum\limits_{i=1}^{C}{\sum\limits_{j=1}^{{{N}_{i}}}{\sum\limits_{k=1}^{K}{\left[ {{d}_{i}}_{jk}+\frac{d_{ijk}^{2}}{4{{\gamma }_{ij}}}+{{\eta }^{2}}{{\gamma }_{ij}}-\eta {{d}_{i}}_{jk} \right]}}}. 
\end{split}\end{equation}

Then substitute (42) into (47), we have
\begin{equation}\begin{split}
& \arg \underset{{{\gamma }_{ij}}}{\mathop{\min }}\,\sum\limits_{i=1}^{C}{\sum\limits_{j=1}^{{{N}_{i}}}{\sum\limits_{k=1}^{K}{\left[ {{d}_{i}}_{jk}+\frac{d_{ijk}^{2}}{4{{\gamma }_{ij}}}+{{\eta }^{2}}{{\gamma }_{ij}}-\eta {{d}_{i}}_{jk} \right]}}} \\ 
& =\arg \underset{{{\gamma }_{ij}}}{\mathop{\min }}\,\sum\limits_{i=1}^{C}\sum\limits_{j=1}^{{{N}_{i}}}\sum\limits_{k=1}^{K}\Bigg[
			{{d}_{i}}_{jk}+\frac{d_{ijk}^{2}}{4{{\gamma }_{ij}}}+\frac{1}{{{K}^{2}}}(1+\frac{d_{ijk}^{2}}{4{{\gamma }_{ij}}^{2}}+\\
			&
			~~~\frac{{{d}_{i}}_{jk}}{{{\gamma }_{ij}}}){{\gamma }_{ij}}  
			 -\frac{1}{K}(1+\frac{{{d}_{i}}_{jk}}{2{{\gamma }_{ij}}}){{d}_{i}}_{jk} 
			\Bigg]. 
\end{split}\end{equation}
Define ${{q}_{ij}}\triangleq \sum\nolimits_{k=1}^{K}{d_{ijk}^{2}}$ and omit the terms irrelevant to ${{\gamma }_{ij}}$, then the problem of (48) can formulated as:
\begin{equation}\begin{split}
& \arg \underset{{{\gamma }_{ij}}}{\mathop{\min }}\,\sum\limits_{i=1}^{C}{\sum\limits_{j=1}^{{{N}_{i}}}{\sum\limits_{k=1}^{K}{\left[ \frac{1}{{{\gamma }_{ij}}}\frac{{{q}_{ij}}}{4}{{\left( 1-\frac{1}{K} \right)}^{2}}+\frac{1}{{{K}^{2}}}{{\gamma }_{ij}} \right]}}} \\ 
& =\arg \underset{{{\gamma }_{ij}}}{\mathop{\min }}\,\sum\limits_{i=1}^{C}{\sum\limits_{j=1}^{{{N}_{i}}}{\sum\limits_{k=1}^{K}{\left[ \frac{a}{{{\gamma }_{ij}}}+b{{\gamma }_{ij}} \right]}}} \\ 
\end{split}\end{equation}
where $a=\dfrac{{{q}_{ij}}}{4}{{\left( 1-\dfrac{1}{K} \right)}^{2}}$ and $b=\dfrac{1}{{{K}^{2}}}$.

Because both the low bound and high bound of ${{\gamma }_{ij}}$ non-negative, according to the inequality of arithmetic and geometric means, the objective function of (49) is bounded: 
\begin{equation}
\sum\limits_{i=1}^{C}{\sum\limits_{j=1}^{{{N}_{i}}}{\sum\limits_{k=1}^{K}{\left[ \frac{a}{{{\gamma }_{ij}}}+b{{\gamma }_{ij}} \right]}}}\ge 2\sum\limits_{i=1}^{C}{\sum\limits_{j=1}^{{{N}_{i}}}{\sum\limits_{k=1}^{K}{ab}}}
\end{equation}
Note that $\sum\limits_{i=1}^{C}{\sum\limits_{j=1}^{{{N}_{i}}}{\sum\limits_{k=1}^{K}{\left[ \dfrac{a}{{{\gamma }_{ij}}}+b{{\gamma }_{ij}} \right]}}}=2\sum\limits_{i=1}^{C}{\sum\limits_{j=1}^{{{N}_{i}}}{\sum\limits_{k=1}^{K}{ab}}}$ holds if and only if 
\begin{equation}
\frac{a}{{{\gamma }_{ij}}}=b{{\gamma }_{ij}}
\end{equation}
holds. Eq. (51) implies the optimal value of ${{\gamma }_{ij}}$ is 
\begin{equation}\begin{split}
 \gamma _{ij}^{*}&=\sqrt{a/b\vphantom{\sum\nolimits_{k=1}^{K}{d_{ijk}^{2}}}}=\sqrt{{{K}^{2}}\frac{{{q}_{ij}}}{4}{{\left( 1-\frac{1}{K} \right)}^{2}}} \\ 
& =\frac{1}{2}\left( K-1 \right)\sqrt{{{q}_{ij}}\vphantom{\sum\nolimits_{k=1}^{K}{d_{ijk}^{2}}}} \\ 
& =\frac{1}{2}\left( K-1 \right)\sqrt{\sum\nolimits_{k=1}^{K}{d_{ijk}^{2}}}. \\ 
\end{split}\end{equation}

Eq. (52) implies that the regularization parameter is related to the sum of the squared distances of neighbors in low-dimensional space. The regularization parameter increases with the distances in low-dimensional space. If the sum of the low-dimensional distances of the neighbors is large, it will give large penalty on the similarity. Therefore, in our method, the regularization parameter is adaptive to the neighbors in low-dimensional space and has intuitive meaning.

\subsubsection{$\mathbf{W}$-step (Compute $\mathbf{W}$ when $\mathbf{S}$ and $\mathbf{R}$ are fixed)}

The goal of  $\mathbf{W}$-step is to learn optimal projection matrix ${{\mathbf{W}}^{*}}$ when the similarity tensor $\mathbf{S}$ and regularization $\mathbf{R}$ are fixed. The corresponding optimiza-tion problem becomes
\begin{equation}
\medmuskip -2mu\thickmuskip -2mu\thinmuskip -2mu
\begin{split}
{{\mathbf{W}}^{*}}&=\arg \underset{\begin{smallmatrix} 
	\mathbf{W} \\ 
	{{\mathbf{W}}^{T}}{{\mathbf{S}}_{t}}\mathbf{W}=\mathbf{I} 
	\end{smallmatrix}}{\mathop{\min }}\,\sum\limits_{i=1}^{C}{\sum\limits_{j=1}^{{{N}_{i}}}{\sum\limits_{k=1}^{{{N}_{i}}}{({{s}_{ijk}}||{{\mathbf{W}}^{T}}{{\mathbf{x}}_{ij}}-{{\mathbf{W}}^{T}}{{\mathbf{x}}_{ik}}||_{2}^{2}+{{\gamma }_{ij}}s_{ijk}^{2})}}} \\ 
&=\arg \underset{\begin{smallmatrix} 
	\mathbf{W} \\ 
	{{\mathbf{W}}^{T}}{{\mathbf{S}}_{t}}\mathbf{W}=\mathbf{I} 
	\end{smallmatrix}}{\mathop{\min }}\,\sum\limits_{i=1}^{C}{\sum\limits_{j=1}^{{{N}_{i}}}{\sum\limits_{k=1}^{{{N}_{i}}}{{{s}_{ijk}}||{{\mathbf{W}}^{T}}{{\mathbf{x}}_{ij}}-{{\mathbf{W}}^{T}}{{\mathbf{x}}_{ik}}||_{2}^{2}}}} \\ 
\end{split}\end{equation}

The minimization problem (53) can be regarded as supervised LPP or LFDA and thus can be formulated as an eigen-decomposition problem. Let $\mathbf{D}\in {{\mathbb{R}}^{N\times N}}$ be a diagonal matrix with its $ii$-entry being ${{D}_{ii}}$.
\begin{equation}
{{D}_{ii}}=\sum\limits_{k=1}^{M}{{{s}_{ijk}}}.
\end{equation} 
The corresponding Laplacian matrix is
\begin{equation}
\mathbf{L}=\mathbf{S}-\mathbf{D}.
\end{equation} 
The optimization problem (53) is then equivalent to
\begin{equation}
\underset{\begin{smallmatrix} 
	\mathbf{W} \\ 
	{{\mathbf{W}}^{T}}{{\mathbf{S}}_{t}}\mathbf{W}=\mathbf{I} 
	\end{smallmatrix}}{\mathop{\min }}\,tr\left( {{\mathbf{W}}^{T}}{{\mathbf{X}}^{T}}\mathbf{LXW} \right).
\end{equation}
where ``\textsl{tr}'' stands for the trace operator. Consequently, the basis vectors ${{\mathbf{w}}_{i}}$ (columns of $\mathbf{W}$) are the eigen-vectors of the  following generalized eigen-decomposition problem:
\begin{equation}
\mathbf{L}{{\mathbf{w}}_{i}}=\lambda {{\mathbf{w}}_{i}}.
\end{equation}

\subsubsection{The Complete Training Algorithm}

Iterations of the $\mathbf{S}$-step, $\mathbf{R}$-step, and $\mathbf{W}$-step form the training algorithm given in Algorithm 1.

\begin{algorithm}
	\caption{The training algorithm of the proposed SLNP method}
	\label{alg1}
	\begin{algorithmic}
		\REQUIRE $C$ classes of $N$ training samples $\mathbf{X}=\{{{\mathbf{x}}_{11}}, {{\mathbf{x}}_{12}}, \ldots,$ $ {{\mathbf{x}}_{1{{N}_{1}}}}, {{\mathbf{x}}_{21}}, {{\mathbf{x}}_{22}}, \ldots,{{\mathbf{x}}_{2{{N}_{2}}}}, \ldots , {{\mathbf{x}}_{C{{N}_{C}}}}\}$ with ${{\mathbf{x}}_{ij}}\in {{\mathbb{R}}^{D\times 1}}$. The number $K$ of neighbors. The number $P$ of iterations.
		\ENSURE Projection matrix $\mathbf{W}\in {{\mathbb{R}}^{D\times d}}$, similarity tensor $\mathbf{S}\in {{\mathbb{R}}^{C\times N\times N}}$, regularization matrix $\mathbf{R}\in {{\mathbb{R}}^{C\times {{N}_{i}}}}$
		
		\textbf{Initialization}:  Initialize $\mathbf{S}$.
		
		\textbf{Iteration:}	
		\FOR {$p=1:P$}
		
		\FOR {$c=1:C$ (for each class)}
		\STATE 1:$\mathbf{W}$-step. 
		
		Compute the diagonal matrix $\mathbf{D}$ by ${{D}_{ii}}=\sum\nolimits_{k=1}^{M}{{{s}_{ijk}}}$. 
		
		Computer by Laplacian matrix by $\mathbf{L}=\mathbf{S}-\mathbf{D}$.
		
		Compute the columns ${{\mathbf{w}}_{i}}$ of $\mathbf{W}$ by eigen-decomposition $\mathbf{L}{{\mathbf{w}}_{i}}=\lambda {{\mathbf{w}}_{i}}$, $i=1,\ldots ,d$.
		\STATE 2:$\mathbf{R}$-step. 
		
		${{\gamma }_{ij}}=\dfrac{1}{2}\left( K-1 \right)\sqrt{\sum\nolimits_{k=1}^{K}{d_{ijk}^{2}}}$.\baselineskip=1.5\baselineskip		
		\STATE 3:$\mathbf{S}$-step
		
		$\eta =\dfrac{1}{K}\left( \dfrac{1}{2{{\gamma }_{ij}}}\sum\limits_{k=1}^{K}{{{d}_{ijk}}}+1 \right)$	\baselineskip=1.1\baselineskip	
		
		${{\mathbf{s}}_{ij}}=-\dfrac{{{\mathbf{d}}_{ij}}}{2{{\gamma }_{ij}}}\text{+}\eta $.	\baselineskip=1.35\baselineskip	
		
		\ENDFOR
		
		\ENDFOR
	\end{algorithmic}
\end{algorithm}

\section{Experimental Results}

We call the proposed method SLNP (Simultaneously Learning Neighborship and Projection Matrix). The training algorithm for the optimal projection matrix $\mathbf{W}$ is given in Algorithm 1. In the test stage, low-dimensional representation $\mathbf{y}$ of a test sample $\mathbf{x}$ is obtained by $\mathbf{y}={{\mathbf{W}}^{T}}\mathbf{x}$. Classifiers can be trained from the low-dimensional version of the training samples. Any type of classifiers can be adopted. Because the emphasis is on the contribution dimensionality, the classical classifier of the nearest neighbor is employed for evaluation of the proposed SLNP method. 

Experiments are conducted on the Extended Yale Face Database B (Yale B) \cite{4,9}, the COIL-100 object dataset \cite{5}, and the MNIST hand-written digits dataset \cite{6}. The proposed method is compared with LDA (PCA+LDA) \cite{8}, FLDA (Fisher Local Discriminant Analysis) \cite{3}, MFA (Marginal Fisher Analysis) \cite{1}, LSDA (Locality Sensitive Discriminant Analysis) \cite{7}.

It is noted that many variants of the above-mentioned methods have been proposed. Despite their success, these methods do not break through the basic frameworks of the classical LDA, FLDA, MFA, and LSDA in the sense of finding neighbors and computing the similarities in the original high-dimensional space. 

Note also that almost all methods employ PCA (Principal Component Analysis) to pre-reduce the dimension of the high-dimensional data in order to avoid the singularity problem or to speed up the training process. Our method also follows the strategy. Let ${{\mathbf{W}}_{PCA}}\in {{\mathbb{R}}^{D\times {{D}_{PCA}}}}$ be projection matrix of PCA. Let ${{\mathbf{W}}_{SLNP}}\in {{\mathbb{R}}^{{{D}_{PCA}}\times d}}$ be projection matrix of SLNP learning from the transformed samples $\mathbf{W}_{PCA}^{T}{{\mathbf{x}}_{ij}}$. The final projection matrix is $\mathbf{W}={{\mathbf{W}}_{PCA}}{{\mathbf{W}}_{SLNP}}$. The number ${{D}_{PCA}}$ of features extracted by PCA is relatively large and the number ${{D}_{SLNP}}$ of features extracted from the PCA features is relatively small. The parameters of ${{D}_{PCA}}$ and  ${{D}_{SLNP}}$ are experimentally determined.

\subsection{Experimental Results on the Extended Yale Face Database B}

The extended Yale Face Database B contains 16,128 images of 28 human subjects under 9 poses and 64 illumination conditions \cite{9}. Examples of the face images are shown in Fig. 3. In our experiments, the image size is normalized to $48\times 42$ pixels. That is, the original image is in $D=48\times 42=2016$ dimensional space.

\begin{figure}[!htb]
	\centering
	\includegraphics[width=3.4in]{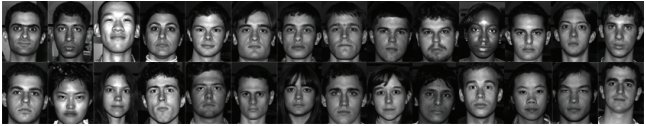}
	\caption{Examples of the face images of the Extended Yale Face Database B.}
	\label{fig3}
\end{figure}

In this section, the convergence of the proposed method and the properties of the learned similarity and regularization parameter are visualized and then the comparison with other methods is described.

\subsubsection{Convergence and the Properties of the Learned Similarity, the Regularization Parameter, and the number of neighbors}~ 

\textbf{Convergence}. To investigate the convergence of the proposed, 10 images per subject are randomly selected as training images. Let ${{D}_{PCA}}=180$, $d={{D}_{SLNP}}=38$, and $K=5$.  
Fig. 4 shows how the objective function $J(\mathbf{S},\mathbf{W},\mathbf{R})$ (see Eq. (19)) varies with iteration number \#. One can see that convergence is achieved when the iteration number is 10. Therefore, the proposed method has good convergence property. 

\begin{figure}[!htb]
	\centering
	\includegraphics[width=3.0in]{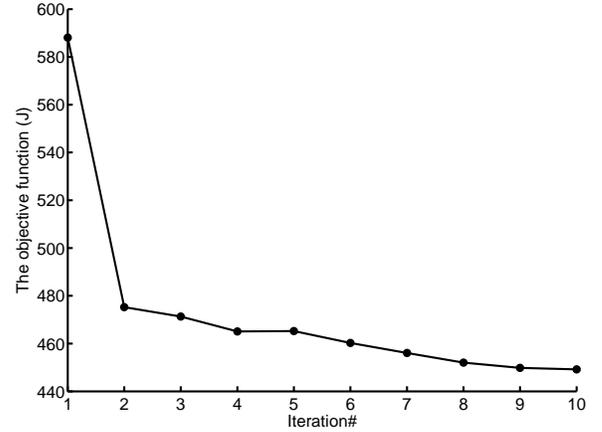}
	\caption{The objective function $J$ as iteration proceeds. }
	\label{fig4}
\end{figure}

\textbf{Property of the Learned Similarities}. Fig. 5 shows how the similarities change with iteration. In Fig. 5, all the face images belong to the same class and class label is 38. The face image in the red rectangle is denoted by the vector ${{\mathbf{x}}_{38,1}}$. From left to right, the rest face images are denoted by ${{\mathbf{x}}_{38,j}}$, $j=2,\ldots,10$.  In the 38-dimensional subspace, the similarities ${{s}_{38,1,k}}$ between the first sample and the rest 9 samples are computed. Prior to iteration, the similarities are equal to 0.1. As iteration proceeds, the similarities change. One see that after the last iteration the similarity between ${{s}_{38,1}}$ and ${{x}_{38,4}}$ is ${{s}_{38,1,4}}=0.1307$  and it is the largest similarity among the similarities between ${{\mathbf{x}}_{38,1}}$ and all the other samples (i.e., ${{\mathbf{x}}_{38,k}}$, $k\ne 1,4$). Moreover, after the last iteration the similarity between ${{\mathbf{x}}_{38,1}}$ and ${{\mathbf{x}}_{38,3}}$ is ${{s}_{38,1,3}}=0.0891$ and it is the smallest similarity among the similarities between ${{\mathbf{x}}_{38,1}}$ and all the other samples (i.e., ${{\mathbf{x}}_{38,k}}$, $k\ne 1,3$). Comparing the images ${{\mathbf{x}}_{38,1}}$,  ${{\mathbf{x}}_{38,4}}$, and ${{\mathbf{x}}_{38,3}}$, we can see that the image ${{\mathbf{x}}_{38,4}}$ has the most similar illumination condition to the image ${{s}_{38,1}}$ whereas the image ${{\mathbf{x}}_{38,3}}$ is quite different from ${{\mathbf{x}}_{38,1}}$. The computed similarities are consistent to our intuition. In summary, the following two phenomena can be observed. (1) The image with the most similar appearance has the largest similarity to reference image and the image with quite different appearance has the smallest similarity to the reference image. (2) Though the similarities between the reference image and all the training images are different, the different is not very large because they belong to the same class. 

\begin{figure}[!htb]
\centering
	\includegraphics[width=3.5in]{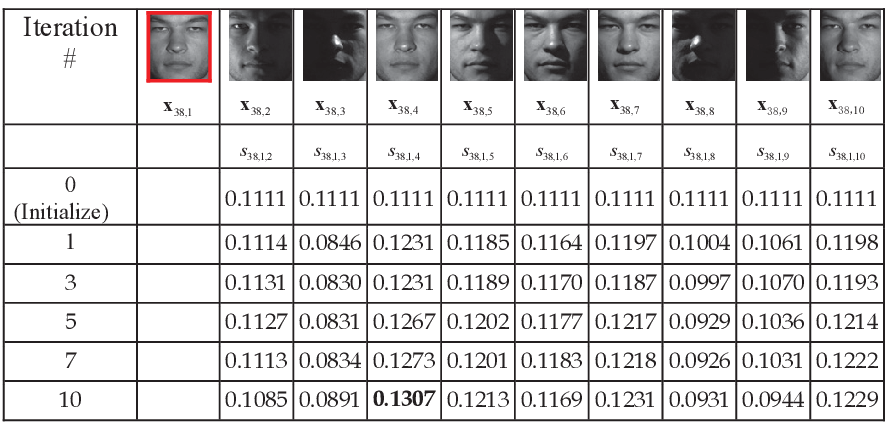}
	\caption{The similarities ${{s}_{38,1,k}}$ ,$k=1, \ldots,9$, changes with iteration.}
	\label{fig5}
\end{figure}

Now we compare the learned similarity and the traditional similarity (i.e., Eq. (13)) used in LPP. The similarity in our method is closely related to the projection matrix and the similarity in LPP is irrelevant to the projection matrix. The face vector ${{\mathbf{x}}_{38,1}}$ is taken as the reference. The other 9 face vectors (i.e., ${{\mathbf{x}}_{38,2}}$, ${{\mathbf{x}}_{38,3}}$, \ldots, ${{\mathbf{x}}_{38,10}}$) are decently sorted according to the similarities between them and the reference ${{\mathbf{x}}_{38,1}}$. Fig. 6(a) shows the sorted results where the proposed similarities are employed and Fig. 6(b) shows results corresponding to the traditional similarities. It is observed that our method is able to give more reasonable sorting results. For example, ${{\mathbf{x}}_{38,1}}$ is most similar to ${{\mathbf{x}}_{38,4}}$ in Fig. 6(a) and is most similar to  ${{\mathbf{x}}_{38,7}}$ in Fig. 6(b). Both ${{\mathbf{x}}_{38,1}}$ and ${{\mathbf{x}}_{38,4}}$ do not have attached shadow below the nose whereas attached shadow exists in ${{\mathbf{x}}_{38,7}}$. Because our similarity is optimal in low-dimensional space, our method is capable of filling the semantic gap. If traditional similarities are employed, the following two phenomena can be observed. (1) The traditional similarity is inferior to the proposed one in the sense of capturing semantic similarity. (2) Though the images belong to the same class, their difference in similarity is very large. For example, ${{s}_{38,1,7}}=0.5063$ whereas ${{s}_{38,1,6}}=0.0061$.

\begin{figure}[!htb]
	\includegraphics[width=3.5in]{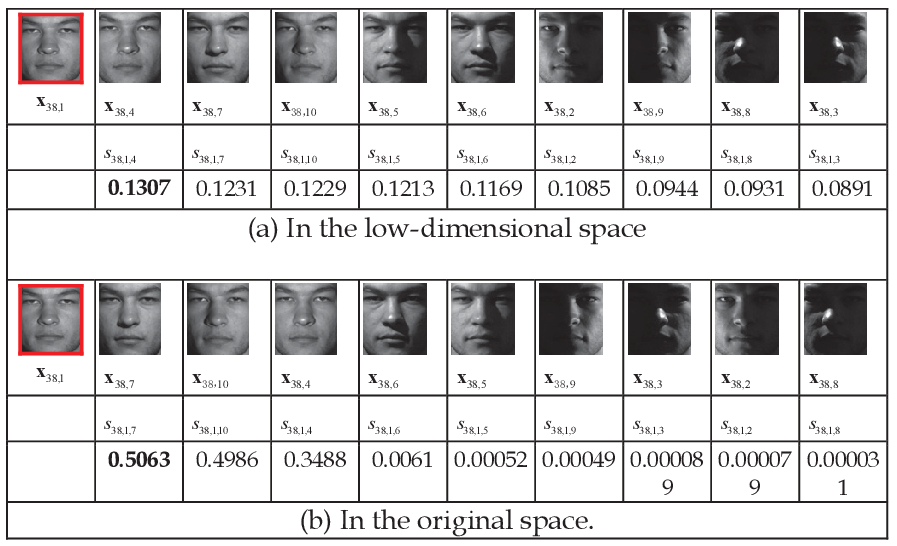}
	\caption{Comparison of the order of similarity in original space (a) and the low-dimensional space (b).}
	\label{fig6}
\end{figure}

\textbf{Property of the Learned Regularization Parameter}. Eq. (52) tells that the regularization parameter is a function of the sum of distances of neighbors in low-dimensional space. The optimal regularization parameter $\gamma _{ij}^{*}$ is obtained by iteratively applying $\gamma _{ij}^{*}=\dfrac{1}{2}\left( K-1 \right)\sqrt{\sum\nolimits_{k=1}^{K}{d_{ijk}^{2}}}$ (i.e., (52)). Different sample $j$ of class $i$ corresponds to different regularization parameter $\gamma _{ij}^{*}$. To intuitively understand the regularization parameter, we compute the average regularization parameter $\gamma _{i}^{*}$ for class $i$:
\begin{equation}
\gamma _{i}^{*}\text{=}\frac{1}{{{N}_{i}}}\sum\limits_{j=1}^{{{N}_{i}}}{\gamma _{ij}^{*}}
\end{equation}

Fig. 7 shows the images of class 27, class 23, and class 24. The corresponding average regularization parameters are $\gamma _{27}^{*}=40.45$, $\gamma _{23}^{*}=20.17$, and $\gamma _{24}^{*}=10.02$, respectively. The order of the average regularization parameters is $\gamma _{24}^{*}<\gamma _{23}^{*}<\gamma _{27}^{*}$. The order relationship can be explained as follows. The intrinsic variation in class 27 is the largest and the variation in class 24 is the least. The regularization parameter is sensitive to the intrinsic variations of the samples. 

\begin{figure}[!htb]
	\includegraphics[width=3.5in]{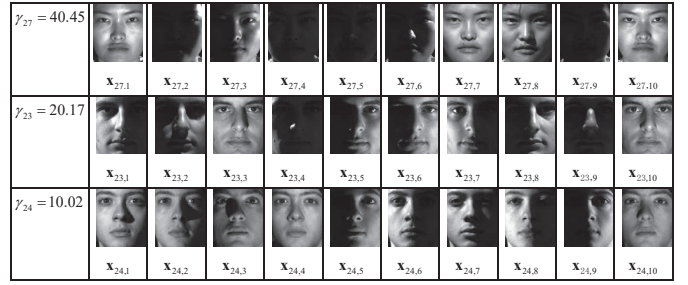}
	\caption{The average regularization parameter $\gamma _{i}^{*}$ reflects the variations in low-dimensional space.}
	\label{fig7}
\end{figure}

In above experiments, the setup of the parameters is: ${{D}_{PCA}}=180$,  ${{D}_{SLNP}}=38$, $K=9$.  

\textbf{Robustness to the number of the neighbors}. From Algorithm 1, one can see that the parameters ${{\gamma }_{ij}}$, $\eta $, ${{s}_{ijk}}$, and $\mathbf{W}$ are learned automatically whereas the number $K$ of neighbors is manually set. Therefore, it is worth investigating whether or not the proposed method is sensitive to the number $K$ of neighbors.
 
Let  $K$ varies from 2 to 9 and compute the recognition rate for each  $K$. Fig. 8 shows the curves of recognition rate versus $K$. It can be seen that the recognition rate is robust to $K$. In the following experiments, we let $K=5$. 

\begin{figure}[!htb]
	\centering
	\includegraphics[width=3.0in]{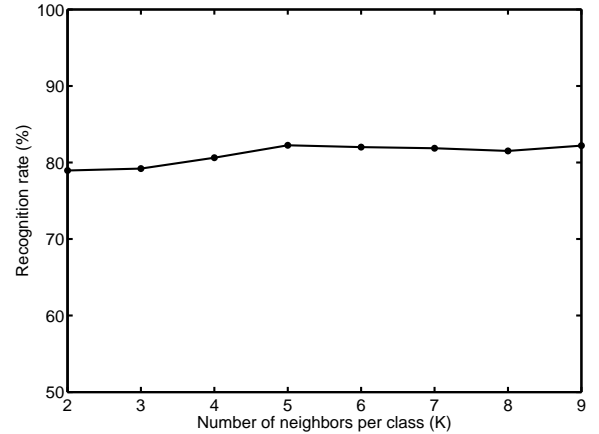}
	\caption{Recognition rate versus the number of nearest neighbors.}
	\label{fig8}
\end{figure}

\subsubsection{Comparison with Other Methods}

${{D}_{PCA}}$ and  ${{D}_{SLNP}}$ (i.e.,  $d$) are experimentally determined. When ${{N}_{i}}=10$ (i.e., 10 images of each class is randomly selected for training and the rest images are used for test) and  ${{D}_{PCA}}=180$, we plot in Fig. 9 how the recognition rate changes with $d$ (i.e., the dimension of final dimension). One can find that the recognition rate increases fast with $d$ until $d=38$ and then become stable and slightly decreases with $d$. So $d=38$ is used for our method to compare with other methods. 

\begin{figure}[!htb]
	\centering
	\includegraphics[width=3.0in]{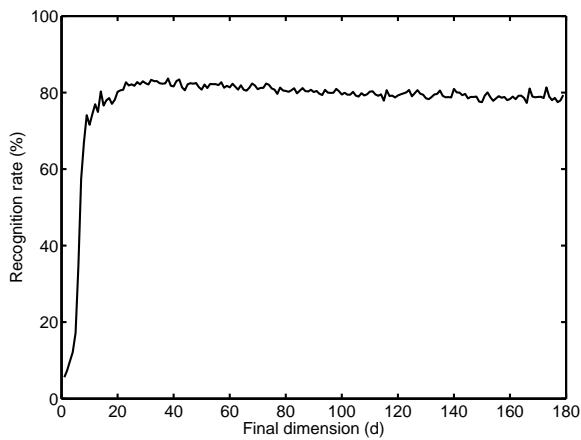}
	\caption{Recognition rate (\%) versus the final dimension $d$.}
	\label{fig9}
\end{figure}

The values of  ${{D}_{PCA}}$ and ${{D}_{SLNP}}$ for different ${{N}_{i}}$ are given in Table \uppercase\expandafter{\romannumeral1}. 

\begin{table}[!htb]
	\caption{${{D}_{PCA}}$ and  ${{D}_{SLNP}}$ for different ${{N}_{i}}$ on the extended Yale database B.}
	\label{table1}
	\centering
	\begin{tabular}{ccccccc}
		\toprule
		${{N}_{i}}$ & 5 & 10 & 15 & 20 & 25 & 30 \\
		\midrule
		${{D}_{PCA}}$ & 110 & 180 & 220 & 400 & 420 & 460 \\
		${{D}_{SLNP}}$ & 38 & 38 & 40 & 38 & 38 & 38 \\
		\bottomrule
	\end{tabular}
\end{table}

Table \uppercase\expandafter{\romannumeral2} gives the recognition rates of different methods when different number of samples per class is for training. One can see that the recognition rates increase with ${{N}_{i}}$. Importantly, for each ${{N}_{i}}$, the proposed SLNP method achieves the best performance. 

\begin{table}[!htb]
	\caption{Comparison in terms of recognition rate (\%) on the extended Yale database B.}
	\label{table2}
	\centering
	\begin{tabular}{ccccccc}
		\toprule
		${{N}_{i}}$ & 5 & 10 & 15 & 20 & 25 & 30 \\
		\midrule
		LDA & 64.55	&80.65	&86.51	&89.89	&91.61 &92.96\\
		MFA & 51.23	&66.67 &70.36	&70.74	&71.71 &73.96\\
		LSDA & 21.30 &48.89	&62.11	&71.58	&77.86 &81.39\\
		LFDA & 61.88 &78.43	&83.79	&86.77	&88.89 &90.02\\
		Ours (SLNP) & \textbf{70.14} &\textbf{84.02} &\textbf{88.50}	&\textbf{93.71}	&\textbf{95.29} &\textbf{96.62}\\
		\bottomrule
	\end{tabular}
\end{table}

\subsection{Experimental Results on the COIL-100 Database}

The COIL-100 database consists of 100 objects (classes) with 72 images per class \cite{5}. The objects were placed on a motorized turntable, which was rotated through 360 degrees at every 5 degrees a time. In our experiment, each image is down-sampled to the size of 16$\times$16 pixels. Examples of the images are shown in Fig. 10.

\begin{figure}[!htb]
	\centering
	\includegraphics[width=3.4in]{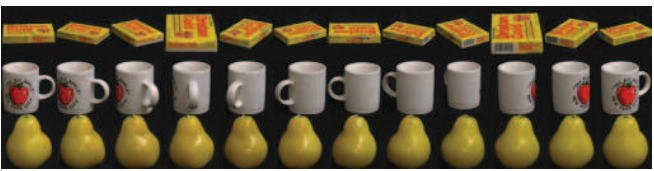}
	\caption{Examples of the COIL-100 database.}
	\label{fig10}
\end{figure}

The number $K$ of neighbors is set to 3. Different number  ${{N}_{i}}$ (i.e., $M$) of samples in each class is used for training and the rest samples are used for test. The parameters ${{D}_{PCA}}$ and  ${{D}_{SLNP}}$ (${{D}_{SLNP}}=d$) corresponding to different ${{N}_{i}}$ are given in Table \uppercase\expandafter{\romannumeral3}.

\begin{table}[!htb]
	\caption{${{D}_{PCA}}$ and  ${{D}_{SLNP}}$ for different ${{N}_{i}}$ on the COIL-100 database.}
	\label{table3}
	\centering
	\begin{tabular}{cccccc}
		\toprule
		${{N}_{i}}$ & 6 & 12 & 18 & 24 & 30 \\
		\midrule
		${{D}_{PCA}}$ & 30 & 32 & 30 & 32 & 40 \\
		${{D}_{SLNP}}$ & 13 & 14 & 17 & 14 & 14 \\
		\bottomrule
	\end{tabular}
\end{table}

The recognition rates of the proposed SLNP method, LDA, MFA, LSDA, and LFDA are given in Table \uppercase\expandafter{\romannumeral4}. One can find that the proposed SLNP achieves the highest recognition rates for all the cases. The superiority of SLNP is remarkable when ${{N}_{i}}$ is 6. In this situation, the recognition rates of SLNP is 85.89\% whereas the recognition rates of LDA, MFA, LSDA, and LFDA are 78.20\%, 76.28\%, 76.03\%, 76.03\%, and 81.14\%, respectively. LSNP outperforms LDA, MFA, LSDA, and LFDA by 7.69\%, 9.61\%, 9.85\%, and 4.75\%, respectively. 

\begin{table}[!htb]
	\caption{Comparison in terms of recognition rate (\%) on the COIL-100 database.}
	\label{table4}
	\centering
	\begin{tabular}{cccccc}
		\toprule
		${{N}_{i}}$ & 6 & 12 & 18 & 24 & 30 \\
		\midrule
		LDA & 78.20	&87.99 &92.40	&94.57	&95.92\\
		MFA & 76.28	&84.22 &87.72	&89.61	&91.04\\
		LSDA & 76.03 &88.30	&92.82	&95.09	&96.43\\
		LFDA & 81.14 &89.45	&92.92	&95.05	&96.35\\
		Ours (SLNP) & \textbf{85.89} &\textbf{92.77} &\textbf{95.83}	&\textbf{97.48}	&\textbf{98.48}\\
		\bottomrule
	\end{tabular}
\end{table}

\subsection{Experimental Results on the MNIST Database}

The MNIST database consists of images of handwritten digits \cite{6}. Fig. 11 shows some examples of the database. 

\begin{figure}[!htb]
	\centering
	\includegraphics[width=3.0in]{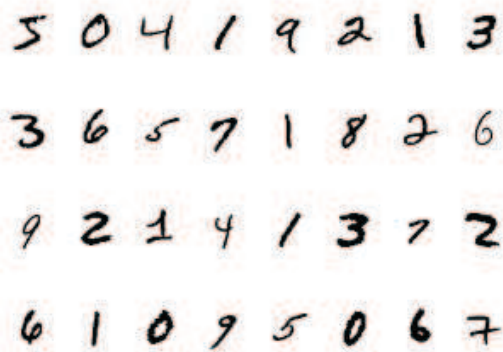}
	\caption{Examples of the MNIST dataset.}
	\label{fig11}
\end{figure}

We randomly samples 6000 images from the dataset. The images are normalized to 14$\times$14 pixels. We first investigate in Fig. 12 how the recognition rate changes with the number $K$ of neighbors when  ${{D}_{PCA}}=32$,  ${{D}_{SLNP}}=d=18$, and ${{N}_{i}}=M=10$. Specially, $K=6$ results in the best recognition performance. Therefore, $K=6$ is used for the following experiments. However, it should be noted that the differences in recognition rates when $K=4, 5, 6, 7,$ and $8$ are not significant. That is, the performance is insensitive to $K$. 

\begin{figure}[!htb]
	\centering
	\includegraphics[width=3.0in]{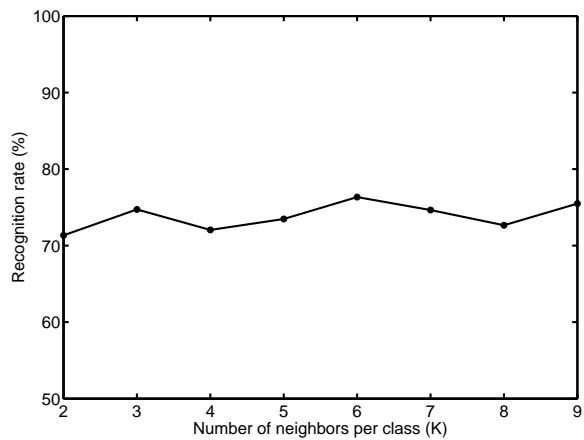}
	\caption{Recognition rate varies with $K$ on the MNIST dataset. }
	\label{fig12}
\end{figure}

Table \uppercase\expandafter{\romannumeral5} shows the values of  ${{D}_{PCA}}$ and ${{D}_{SLNP}}=d$ for different ${{N}_{i}}$. 

\begin{table}[!htb]
	\caption{${{D}_{PCA}}$ and  ${{D}_{SLNP}}$ for different ${{N}_{i}}$ on the MNIST dataset.}
	\label{table5}
	\centering
	\begin{tabular}{ccccccc}
		\toprule
		${{N}_{i}}$ & 5 & 10 & 15 & 20 & 25 & 30 \\
		\midrule
		${{D}_{PCA}}$ & 34 & 32 & 32 & 32 & 32 & 26 \\
		${{D}_{SLNP}}$ & 18 & 18 & 26 & 16 & 30 & 21 \\
		\bottomrule
	\end{tabular}
\end{table}

With the above parameters, the recognition rates of the proposed SLNP are given Table \uppercase\expandafter{\romannumeral6} where comparison with other methods is also given. Generally speaking, the advantage of the proposed SLNP over the other methods becomes significant when the number of samples per class is small. When ${{N}_{i}}=M=25$, the recognition rates of SLNP, LFDA, LSDA, MFA, and LDA are 83.17\%, 81.01\%, 80.42\%, 77.64\%, and 77.10\%, respectively. When ${{N}_{i}}=M=5$, the recognition rates of SLNP, LFDA, LSDA, MFA, and LDA are respectively 67.51\%, 60.23\%, 54.95\%, 59.51\%, and 62.22\%. SLNP outperforms LFDA by 7.28\% when ${{N}_{i}}=M=5$ and outperforms LFDA by 2.16\% when ${{N}_{i}}=M=25$. 

\begin{table}[!htb]
	\caption{Comparison in terms of recognition rate (\%) on the MNIST dataset.}
	\label{table6}
	\centering
	\begin{tabular}{ccccccc}
		\toprule
		${{N}_{i}}$ & 5 & 10 & 15 & 20 & 25 & 30 \\
		\midrule
		LDA & 62.22	&70.10	&73.21	&75.16	&77.10 &77.85\\
		MFA & 59.61	&69.38 &72.60	&75.88	&77.64 &78.75\\
		LSDA & 54.95 &69.67	&74.86	&77.76	&80.42 &81.82\\
		LFDA & 60.23 &72.78	&76.72	&79.08	&81.01 &82.11\\
		Ours (SLNP) & \textbf{67.51} &\textbf{76.34} &\textbf{79.88}	&\textbf{82.00}	&\textbf{83.17} &\textbf{84.16}\\
		\bottomrule
	\end{tabular}
\end{table}

\section{Conclusion}

In this paper, we have presented a supervised dimensionality reduction. By letting the similarity and neighbors depend on projection matrix, we have proposed an objective function consists of a similarity data term and a similarity norm penalty term and imposed nonnegative and sum-to-one constraints on the similarity. An alternative algorithm has been developed to the optimal similarities, projection matrix, regularization parameter, Lagrangian multiplier. Theoretical analysis showed that the optimal similarities, regularization parameter, and Lagrangian multiplier are functions of distances in low-dimensional space spanned by the projection matrix. There are almost no parameters to be tuned except the number of neighbors and the algorithm is not sensitive to the number of neighbors.


%

\appendices




\ifCLASSOPTIONcaptionsoff
  \newpage
\fi

\end{document}